\documentclass{llncs}

\usepackage{color,amsmath}
\usepackage{graphicx}
\usepackage{subfig}
\usepackage{listings}
\usepackage[noend]{algpseudocode}
\usepackage{algorithm}
\usepackage{svn-multi}
    \svnidlong
    {$HeadURL: https://xp-dev.com/svn/norm-detect/trunk/coin-norm-detect-2013/coin-norm-detect-main.tex $}
    {$LastChangedDate: 2013-03-08 16:44:32 -0300 (Fri, 08 Mar 2013) $}
    {$LastChangedRevision: 50 $}
    {$LastChangedBy: meneguzzi $}

\let\oldmarginpar\marginpar
\renewcommand\marginpar[1]{\-\oldmarginpar[\raggedleft\footnotesize #1]%
{\raggedright\footnotesize #1}}
\ifx\comment\undefined
\newenvironment{comment}[2][Note]{\footnote{\textbf{#1:} #2}}
\fi

\ifx\todo\undefined
\newcommand\todo[1]{~\newline{\color{red}\framebox[\textwidth]{\parbox{.95\textwidth}{TODO: #1}}}~\newline}
\fi

\newcommand{\Obl}[0]{\ensuremath{\mathbf{O}}}
\newcommand{\Frb}[0]{\ensuremath{\mathbf{F}}}

\title{Norm Identification through Plan Recognition}
\author{
   Nir Oren\inst{1} 
   \and Felipe Meneguzzi\inst{2}
   }
\institute{
Department of Computing Science, University of Aberdeen \\
  Aberdeen, AB24 3UE, Scotland\\
  \email{n.oren@abdn.ac.uk}
\and School of Computer Science, 
Potifical Catholic University of Rio Grande do Sul\\
Porto Alegre, 90619-900, Brazil\\
\email{felipe.meneguzzi@pucrs.br}}

\begin{document}
\maketitle

\begin{abstract}
Societal rules, as exemplified by norms, aim to provide a degree of behavioural stability to multi-agent societies.
Norms regulate a society using the deontic concepts of permissions, obligations and prohibitions to specify what can, must and must not occur in a society. 
Many implementations of normative systems assume various combinations of the following assumptions: that the set of norms is static and defined at design time; that agents joining a society are instantly informed of the complete set of norms;  that the set of agents within a society does not change; and that all agents are aware of the existing norms. 
When any one of these assumptions is dropped, agents need a mechanism to identify the set of norms currently present within a society, or risk unwittingly violating the norms.
In this paper, we develop a norm identification mechanism that uses a combination of parsing-based plan recognition and Hierarchical Task Network (HTN) planning mechanisms, which operates by analysing the actions performed by other agents. While our basic mechanism cannot learn in situations where norm violations take place, we describe an extension which is able to operate in the presence of violations.
\end{abstract}

\section{Introduction}
\label{sec:intro}

Large scale multi-agent societies must be designed with resilience in mind, permitting agents to join or leave at any time. To ensure that the society functions as intended, constraints must be imposed on agent actions. Such constraints often take the form of norms --- soft constraints which, when instantiated, oblige, prohibit or permit an agent to see to it that some state of affairs holds. Norm compliance by agents occurs for a variety of reasons, including rationality, fear of punishment, or benevolence.
Now when a new agent enters the society, it must be made aware of the society's norms. 
When such norms are codified, obtaining them can be easy. 
However, even in such situations, factors such as limited bandwidth could make the transmission of the set of norms difficult. 
Additional difficulties --- which are endemic to open multi-agent systems --- such as  the lack of a shared ontology, or norms being implicitly rather than explicitly specified, mean that a new agent must instead identify the norms with little or no assistance from its designer or other agents in the system. 

One popular approach to norm identification\footnote{In the literature, the task which we refer to as norm identification has also been called norm learning, norm recognition, and norm detection.} through learning \cite{Andrighetto2007,Savarimuthu2009,SavarimuthuJASSS2010}, exemplified by \cite{Savarimuthu2010a}, utilises what is referred to as an \emph{observation-based} technique to track the behaviour of others in order to infer the norms currently in force. 
This technique is based on the detection of a \emph{violation signal}, detecting when another agent has violated a norm, and by identifying the violating situation, learning the associated norm. 
While such an approach works well when norms are regularly violated and sanctions are explicitly applied, learning from violation signals when a system's agents (largely) comply with norms is difficult (if not impossible). 
In this paper, we propose an observational technique for norm learning that operates by observing compliant agents instead. We then extend this base technique to learn from both compliant, and violating agents. 

At the heart of our technique lie two components --- a planner and a plan recogniser.  Using these, an agent entering the system observes the actions of other agents in the system and utilises the plan recogniser to identify their overarching goals.  It then utilises the planner to generate alternative plans that achieve these goals.  By comparing the  plans other agents are actually executing to those generated by the planner, avoided (or repeatedly visited) actions and  states can be identified.  
Repeatedly executing the plan recognition and planning steps over time enables the agent to conclusively identify those actions and states that are always avoided or visited in the execution of a plan. 
Since such states are analogous to prohibited or obliged situations, the agent can use them to identify norms. 
This basic approach cannot easily handle situations where agents \emph{occasionally} violate norms, and we extend our technique to be able to handle norm violations by agents. 

Our technique assumes that agents in the society have access to common \emph{domain knowledge} in the form of a shared plan library that does not change over time, as well as a shared notion of the natural rewards within the system. 
Moreover, agents have no instant knowledge of the norms actually being enforced in the system. 
The reason for this assumption is that the plan library and the natural rewards in the environment are assumed to be based on some aspect of the \emph{physical} structure of the environment. As an example, in the context of a road system, knowledge about routes to move from point A to point B might be known to all agents, as is knowledge that a certain road is full of potholes.  However, the existence of a contextual prohibition on driving fast, exemplified by knowledge that the motorway police are strictly enforcing speed limits at a certain road might not be commonly known. 

The remainder of this paper is structured as follows. The next section  provides the background needed to understand the remainder of the paper, describing our planning formalism, outlining our plan recognition technique, and the nature of norms within the domain. Section~\ref{sec:normRecognition} then describes our approach to norm identification. We examine related and future work in Section \ref{sec:discussion}, before concluding in Section~\ref{sec:conclusion}.

%
%
%
%
\section{Background}
\label{sec:background}

In this section, we review the underlying techniques upon which we build our norm identification mechanism. 
We start by formally describing the planning model we will use to recognise alternate plans agents could execute, and follow this by detailing the plan recognition approach we use to analyse agent behaviour. 
Finally, we review related norm recognition approaches. 

\subsection{HTN Planning}
\label{subsec:htnPlanning}

We model the environment in which agents operate as a state transition system, consisting of individual states, each of which is a set of ground atoms of a first order language $\mathcal{L}$ containing finitely many predicate and constant symbols, as well as an infinite number of variable symbols\footnote{We adopt the prolog convention of representing constants with an initial lowercase letter, and variables with an initial uppercase letter.} and no function symbols. 
This model is taken from the standard model used for planning, as described in \cite{Ghallab2004}.

An atom $p$ holds in a state $s\in S$ iff $p \in s$. Given a set of literals $g$, $s$ satisfies $g$ (written $s \models g$) when there is some substitution $\sigma$ such that every positive literal of $\sigma(g)$ is in $s$, and no negated literal of $\sigma(g)$ is in $s$.

The execution of an \emph{action} by an agent causes the system to transition between states. 
In the planning literature, families of actions are specified using an action template (\emph{operator}) containing variables referring to abstract elements in the domain. 
We specify such operators through a triple
$$o=(\mathit{name(o),pre(o),post(o)})$$
Here, $\mathit{name(o)}$ is a unique name for the operator, $\mathit{pre(o)}$ is a set of literals of $\mathcal{L}$, and $\mathit{post(o)}$ consists of two sets of literals, $\mathit{post^{+}(o)}$ and $\mathit{post^{-}(o)}$. $\mathit{pre(o)}$ identifies the preconditions that must hold in the state in which the operator is executed. $\mathit{post(o)}$ represents the effects, or post conditions, of executing the operator, in terms of literals added (in the case of  $\mathit{post^{+}(o)}$) and removed (in the case of  $\mathit{post^{-}(o)}$). 
The operator's name consists of an operator symbol, followed by a vector of distinct variables, such that all free variables in $\mathit{pre(o),post(o)}$ appear within this vector.

An action is a grounding substitution $\sigma$ over all variables in the operator.  
An operator instance $o$ is \emph{applicable} to a state $s$, $s \models \mathit{pre}(o)$. 
An application of an action to a state $s$ is either a new state $s'$ such that $s'=(s \cup \mathit{post^{+}(o)} \backslash \mathit{post^{-}(o)})$ when $o$ is applicable to $s$, or $s$ otherwise.

In order to achieve a goal, an agent executes a plan --- a sequence of actions $\pi=\langle a_{1}, \ldots, a_{n}\rangle$. 
To identify this sequence, an agent can utilise classical planning techniques \cite{Ghallab2004}. However, classical planning is computationally expensive, increasing the difficulty of creating agents which operate in (near) real-time. 
To overcome this, many agent frameworks have introduced the concept of a \emph{plan library}. 
Such a plan library contains a set of plans generated offline, which can be composed to achieve high-level goals. 
In order to perform this composition, higher level plans are made up of lower level plans, which in turn eventually reduce to primitive tasks that can be directly mapped to actions. 
These plans have a declarative component --- multiple sub-plans might be feasibly invoked to achieve a single step in a high-level plan, and a selection between all these feasible sub-plans must be made. 
We can describe the problem of composing these sub-plans to achieve a high-level goal as a HTN (or rather STN) planning problem.

A hierarchical task network (HTN)  planner \cite{Erol1994} aims to decompose a set of high-level tasks, encoded as a \emph{task network}, into a set of primitive tasks. 
The task network is a directed graph, consisting of the set of tasks, as well as temporal constraints between them, identifying what task must execute before what other tasks. 
We assume that all task networks are acyclic. 
Each task is an expression of the form $$t(r_{1},\ldots, r_{n})$$ Here, $t$ is a unique task symbol and $r_{1}, \ldots r_{n}$ are terms. 
Non-primitive tasks represent high-level tasks (e.g.~$\mathtt{travel(S,D)}$ might represent the task of travelling from some source $\mathtt{S}$ to a destination $\mathtt{D}$). 
Several \emph{methods} can satisfy these tasks, for example to travel between two points one could fly or catch a train. 
We encode a method as a 4-tuple 
$$m=( \mathit{name(m),task(m),precond(m),network(m)})$$ 
$\mathit{name(m)}$ is the name of the method, represented via a unique method name and a vector of terms. 
$\mathit{task(m)}$ encodes the task that the method can refine, while $\mathit{precond(m)}$ consists of a set of positive and negative precondition literals $\mathit{precond}^{+}(m)$, $\mathit{precond}^{-}(m)$, which must be satisfied by the state in order for the method to be \emph{applicable}. 
Finally, the method identifies what tasks must be carried out in order to further refine $\mathit{task(m)}$ represented via a task network.

As an example, the method to fly from a source to a destination might consist of the task network containing tasks requiring one to buy a ticket; go to the airport; and fly to the destination. 
The first and last  tasks could be primitive, while the middle task could be achieved via additional methods (e.g. going by bicycle, by car or by public transport). 
Such compound tasks can therefore be further decomposed by methods, into additional tasks.

Given an initial state, a task network identifying a set of high-level tasks that must be achieved and a set of operators and methods, an HTN planner searches for a plan by finding an appropriate decomposition of tasks (via methods). 
The precise manner in which this decomposition takes place, and its properties, are not critical for this paper, and we refer the reader to \cite{Ghallab2004} for further details. 

It is important to note that an HTN planning problem can be viewed as picking a set of leaf nodes from an AND/OR tree such that if a set of nodes has an AND parent, all nodes are picked, and if a set of nodes has an OR parent, only one node is picked. 
The exact sequence of the leaf nodes then depends on the temporal constraints imposed by the task network. 
Furthermore, each OR node represents a task, and given a plan, it is possible to identify both the set of tasks, and specific task instances (i.e. ground tasks) that form the plan. 
Doing so lies at the heart of plan recognition.

\subsection{Plan Recognition}
\label{subsec:planRecognition}

A large body of work has examined the plan recognition problem \cite{Armentano2007,Geib2007,Geib2009,Sukthankar2007}. 
Due to its simplicity, we focus on a parser based plan recogniser in this paper. 
However, it is important to note that our plan recognition problem is an instance of keyhole plan recognition, and this approach can be trivially interchanged with more complex plan recognisers, such as~\cite{Geib2009}.

Plan recognition aims to identify the actions the agent will follow (including future actions) in order to achieve its goals. 
\cite{Armentano2007} provides a detailed survey of such approaches, noting that in the general case, issues such as partial observability, the interleaving of plans and agent capabilities make the problem a difficult one to address. 
In this paper we ignore many of these complicating factors, and use an NLP-based approach (namely parsing algorithms c.f.~\cite{Geib2007}) to perform plan recognition. 
Such an approach utilises a technique similar to parts of speech tagging in order to identify a plan from within a plan library (or equivalently within a HTN planner). 
Performing plan recognition through parsing arises from the straightforward correspondence between HTN formal structures and those of context-free grammars (CFGs)~\cite{Geib2007}. 
Whereas plan generation for an HTN problem consists of successively refining a task network containing \emph{non-primitive} tasks into one containing only \emph{primitive tasks}, language generation using a CFG consists of transforming an initial string containing \emph{non-terminal symbols} into one containing only \emph{terminal symbols}. 
Refinements in an HTN planning process are made through \emph{methods} which replace a non-primitive task into other, presumably less abstract, tasks that are either primitive or non-primitive. 
Analogously, language generation in CFGs consists of applying \emph{production rules} to strings and replacing non-terminal symbols with other symbols which are either terminal or non-terminal. 
Based on these correspondences, techniques that can reconstruct which production rules were used to generate a certain string in a CFG can be, with some adaptation, be used to determine which methods were used to generate the intermediary non-primitive tasks that resulted in a certain plan from an HTN domain. 

\begin{figure}[ht]
\begin{center}
	\subfloat[Parse Trees]
	{\includegraphics[scale=.5]{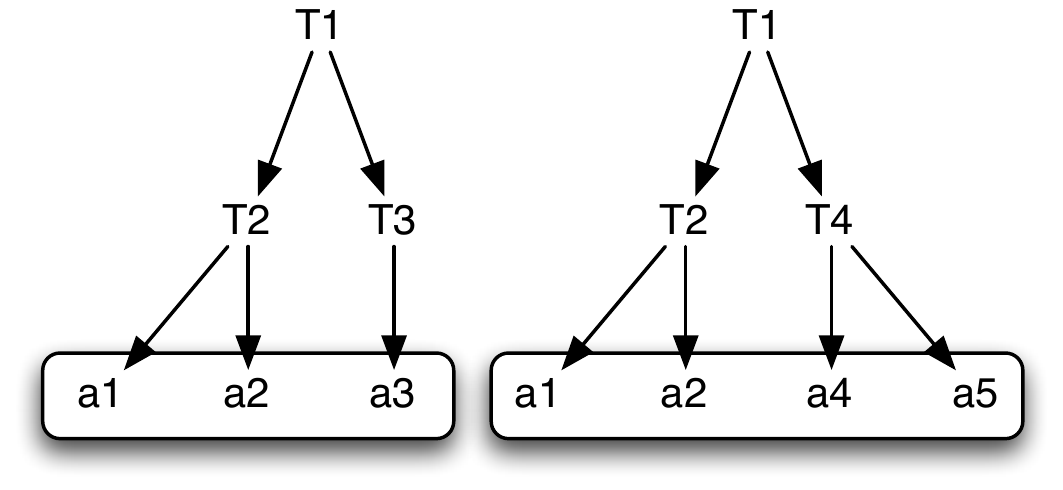}\label{fig:parseTrees}}
	\subfloat[HTN Tree]
	{\includegraphics[scale=.5]{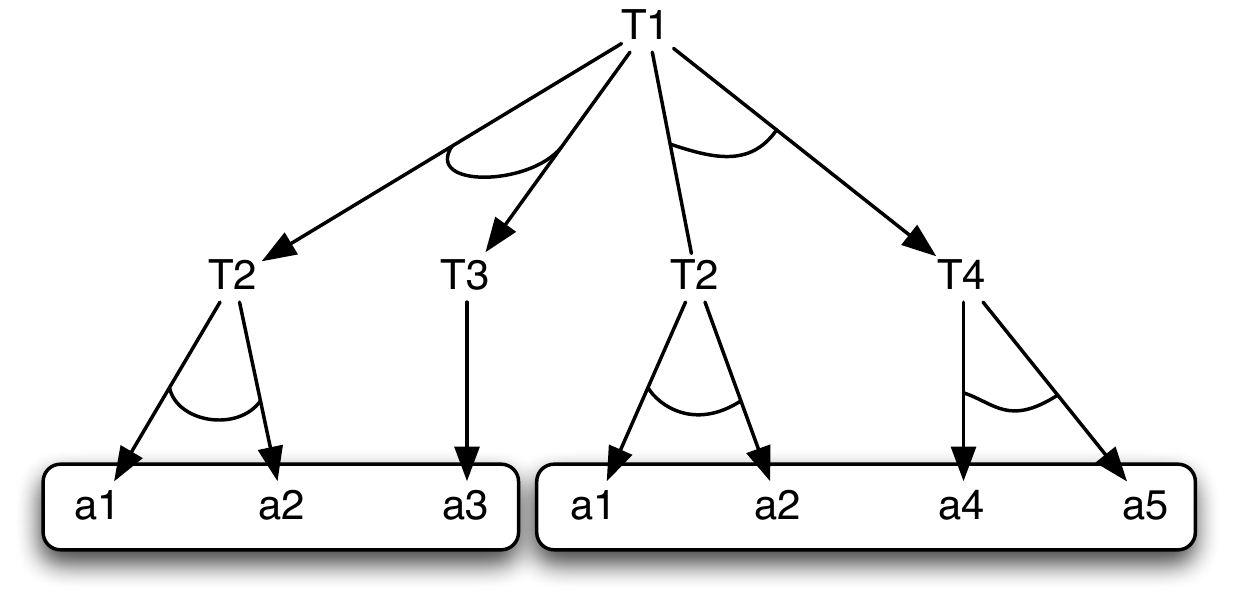}\label{fig:htnTree}}
\end{center}
\caption{HTN and Parse trees}
\label{fig:trees}
\end{figure}

As an example, let us consider a grammar $\langle IS, NS, P, IS\rangle$ with a set of non-terminal symbols $NS = \{ T_1, T_2, T_3, T_4, T_5\}$, a set of terminal symbols $S = \{a_{1},a_{2},a_{3},a_{4},a_{5} \}$, a starting symbol $IS = T_1$, and the following set $P$ of five production rules.
\begin{align*}
	T_1 & \rightarrow T_{2}T_{3} \\
	T_1 & \rightarrow T_{2}T_{4} \\
	T_2 & \rightarrow a_{1}a_{2} \\
	T_3 & \rightarrow a_{3} \\
	T_4 & \rightarrow a_{4}a_{5} \\
\end{align*}
This grammar is capable of generating two different strings `$a_{1}a_{2}a_{3}$' and `$a_{1}a_{2}a_{4}a_{5}$', each of which, when parsed, generates the corresponding parse trees in Figure~\ref{fig:parseTrees}. 
The parse trees \emph{explain} how each string is generated in terms of the production rules used in their derivation, all the way to the initial symbol. 
If we now consider the elements of the sets $S$ and $NS$ to correspond, respectively, to the elements of the sets of primitive and non-primitive tasks of an HTN domain, and convert the production rules of the form $\nu \rightarrow \omega$ to correspond to methods $m=( \mathit{name(m),\nu,\top,\omega})$, then we would arrive at the and-or tree of the potential HTN expansions of task $T_1$ illustrated in Figure~\ref{fig:htnTree}. 
Since the reverse conversion is just as straightforward, we can see how a parsing-based technique can be used to determine what higher-level tasks explain the observation of a particular sequence of actions (a plan). 

\subsection{Norms}
\label{subsec:norms}

The previous section described one portion of our model, namely the domain and the structure of agent plans. 
In this section, we formalise our notion of a norm, which comprises a simplification of the usual elements of a norm tuple~\cite{Aldewereld2006,oren2008}, namely: a \emph{deontic modality}, a \emph{context condition} representing the situations in which a norm should be enforced, and a \emph{normative condition} that, together with the deontic modality, identifies expected behaviour.
We note that whereas the conditions of a norm tuple are normally formalised in terms of a logic formula on the environment state, our norm language uses either fully specified states, or \emph{task} symbols. 

Syntactically, we write a norm as $\mathbf{X}_{y}z$, where $\mathbf{X} \in \{\Obl,\Frb\}$,  $y$ is a task, and $z$ is a task or state. 
$\mathbf{X}$ is the modality of the norm, either an obligation (\Obl) or a prohibition (\Frb). $y$ identifies the norm's context, specifying the situations under which the norm's condition, $z$ must occur (in the case of an obligation), or must not occur (in the case of a prohibition). 
A task $z$ occurs in context $y$ iff $z$ is a subtask of $y$, and under the appropriate grounding of variables, the leaf tasks associated with its OR node are executed. 
In other words, a $z$ occurs in the context of $y$ if, in the process of task $y$'s methods being executed, task $z$ is executed. 
A state $z$ occurs in context $y$ if the state $z$ is entered while the leaf tasks associated with $y$'s OR node are executed. 
In other words, state $z$ is entered while a method whose task is $y$, is being executed. 
If a norm's condition must occur and does not, or alternatively must not occur and does, then the norm is \emph{violated}.

As an example, consider the task of travelling from Aberdeen to Paris, and assume that when travelling anywhere there is a prohibition imposed on transiting via London. 
The task instance could be represented as $\mathtt{travel(aberdeen,paris)}$, with the prohibition represented as $\Frb_{{travel(X,Y)}}\mathit{at(london)}$. 
Now one could have  the method
\begin{align*}
(\mathit{fly(X,Y),travel(X,Y),\{at(X),connect(X,Z),connect(Z,Y)\}}, \\
\{ \mathit{goto(X,Z) \prec goto(Z,Y)}\})
\end{align*}
Here, $\prec$ represents a temporal constraint, stating that $\mathit{goto(X,Z)}$ must occur before $\mathit{goto(Z,Y)}$. Finally, the $\mathit{goto}$ primitive task is represented by the following operator.
$$(\mathit{goto(X,Y)},\{\mathit{at(X)}\},\{\neg \mathit{at(X),at(Y)}\})$$
Given the preconditions $\mathtt{connect(aberdeen,london),connect(london,paris)}$, the instantiation of the plan will result in $\texttt{at(london)}$ occurring in context $\mathtt{travel(aberdeen,paris)}$, violating the norm.
Notice that, although using \emph{tasks} in the context and norm conditions is syntactically different from the state formulas used in traditional norm representations, the net effect of our norm representation is equivalent in that these conditions represent subsets of the domain's state space. 
Given that the state space that \emph{can be} traversed by an agent is completely determined by the tasks in its plan library, this representation is sufficient to encode all the norms that can be detected by our algorithm.


\section{Norm Identification Approaches}
\label{sec:normRecognition}

We are now in a position to describe our main contribution --- a plan recognition based approach to norm identification.
The basic assumption of our approach is that an individual agent's behaviour is defined in terms of a \emph{plan library} of procedural plans following the tradition of agent programming languages such as AgentSpeak(L) \cite{Rao1996}. 
This kind of plan library is analogous to domain specifications for Hierarchical Task Network (HTN) planning. 
Like procedural goals in AgentSpeak(L), goals in HTNs are defined in terms of a partially ordered set of high-level abstract tasks (or \emph{task network}), which can be recursively refined to more concrete tasks until a fully specified plan of executable actions is generated. 
It is straightforward to see that the set of plans an agent can execute is completely defined by such a library, much in the same way in which production rules define all valid strings that could be generated by a grammar. 
Consequently, by being aware of the plan library employed by agents, one can use the sequences of actions executed by these agents to infer the higher-level goals that they pursue. 

We begin this section by describing a novel plan recognition based approach to norm identification. 
This basic approach assumes agents will never violate norms.  
In order to overcome this assumption, we then describe an extension to the basic model.

\subsection{Norm Identification via Plan Recognition}

Our mechanism seeks to identify obligations and prohibitions within the domain, and conceptually operates in several steps. 
As input, we take in a set of \emph{runs}, each representing the states transitioned through by the system in the process of achieving some top level goal. Note that these runs could originate from a single agent pursuing multiple goals over an extended period of time; multiple agents each pursuing goals (though we assume non-interference between agent actions); or a mixture of the two. Algorithm~\ref{alg:learnNorms} formalises our mechanism.

\begin{algorithm}[t]
{\small
\begin{algorithmic}[1]
\Require $R$, a set of runs
\Function{LearnNorms}{$R$}
  \State $\mathit{potO} \leftarrow$ all possible obligations
  \State $\mathit{potF} \leftarrow \emptyset$,  $\mathit{notF} \leftarrow \emptyset$
  \ForAll{$r \in R$} 
   \State $\mathit{pO} = \emptyset$
   \State $\mathit{pF} = \emptyset$ 
   \State $\pi \leftarrow$ \Call{planRecogniser}{$r$} \label{alg:recPlan}

    \ForAll{tasks $t \in \pi$}  
      \ForAll{subtasks $t' \in \pi$ of $t$} \label{alg:phase1S}
        \State $\mathit{pO} \leftarrow \mathit{pO} \cup \{\Obl_{t}t' \}$ \label{alg:addpO}
        \State $\mathit{notF} \leftarrow \mathit{notF} \cup \{ \Frb_{t}t' \}$ \label{alg:addnotF}
      \EndFor 
      \ForAll{state $s$ transitioned through as part of $t$} \label{alg:stateS}
        \State $\mathit{pO} \leftarrow \mathit{pO} \cup \{\Obl_{t}s \}$
        \State $\mathit{notF} \leftarrow \mathit{notF} \cup \{ \Frb_{t}s \}$
      \EndFor \label{alg:stateF}
 \label{alg:phase1F}   
      \ForAll{$\pi' \in$ all possible plans with the same start state as $\pi$ and with a goal identical to $t$} \label{alg:phase2S}   
        \ForAll{tasks $\tau \in \pi' $}
          \ForAll{subtasks $\tau' \in \pi'$ of $\tau$, which are not subtasks of $\tau \in \pi$} \label{alg:addpFTS}
            \State $\mathit{pF} \leftarrow \mathit{pF} \cup \{ \Frb_{\tau}\tau' \}$
          \EndFor \label{alg:addpFTF} 
          \ForAll{states $\Sigma$ visited as part of $\tau \in \pi'$ and not for $\tau \in \pi$} \label{alg:addpFSS}
            \State $\mathit{pF} \leftarrow \mathit{pF} \cup \{\Frb_{\tau}\Sigma \}$
          \EndFor \label{alg:addpFSF} 
        \EndFor 
     \EndFor   \label{alg:phase2F} 
      \EndFor 
    \State $\mathit{potF} \leftarrow (\mathit{potF} \cup \mathit{pF}) \backslash \mathit{notF}$
    \State $\mathit{potO} \leftarrow \mathit{potO} \cap_{t} \mathit{pO}$
  \EndFor 
  \State \Return $\mathit{potO}$, $\mathit{potF}$  
\EndFunction
\end{algorithmic}
}
\caption{The basic norm identification mechanism}
\label{alg:learnNorms}
\end{algorithm}

The algorithm uses several variables: $\mathit{potO}$ stores all potential obligations in the system; $\mathit{potF}$ and $\mathit{notF}$ respectively store potential prohibitions as well as those states of affairs which are definitely not prohibited. When the algorithm starts, $\mathit{potO}$ is initialised with all possible obligations in the system. Obligations are monotonically removed from this set as the algorithm runs. Since we assume a finite number of predicate and constant symbols, $\mathit{potO}$ will be finite, but potentially very large. Within an implementation, techniques such as masking, or storing only those elements not in the set can be used to mitigate the issues surrounding its size. \label{pg:mitigation}

The algorithm operates over a set of runs, identifying norms for each such run. This is done by utilising the plan recogniser to identify the plan $\pi$ being executed (Line \ref{alg:recPlan}). We assume that only one  plan is ever returned by the plan recogniser. This plan $\pi$ recursively identifies high-level tasks and the subtasks used to decompose them into lower level tasks, all the way  down to the primitive task level. Collectively, these tasks, subtasks and primitive tasks are the tasks of the plan, and we refer to any descendent of a task as the task's subtask.

Broadly speaking, our algorithm operates in two phases. Lines \ref{alg:phase1S}--\ref{alg:phase1F} consider only the plan, while Lines \ref{alg:phase2S}--\ref{alg:phase2F} consider alternate plans which achieve the same goals. Examining the first phase in more detail, we iterate over all tasks $t$ of the recognised plan, and over each task's subtasks. Any of these subtasks visited could, potentially, be one that must be visited according to the obligations in the system, and we therefore add it to a temporary variable $\mathit{pO}$ (Line \ref{alg:addpO}). Furthermore, since we assume that no violations can occur, the executed subtask cannot be prohibited in the context of the parent task, and we therefore add it to the list of prohibitions that definitely do not exist in Line \ref{alg:addnotF}. Norms can describe both tasks and states, and an identical procedure to the one followed above is carried out to identify potentially obligatory, and definitely not prohibited states (Lines \ref{alg:stateS}--\ref{alg:stateF}).

Line \ref{alg:phase2S} identifies all other possible plans that began with the same initial state as the actually executing plan, and finish by achieving the same goals as the current task. The aim here, and in the remainder of the second phase, is to identify tasks that could have been, but were not executed, as these represent potential prohibitions that were complied with. To do so, we iterate over all tasks $\tau$ within an alternate plan, and over all subtasks $\tau'$ of  $\tau$, where $\tau'$ is  not a subtask of the $t$ (the task within the original plan). Thus, each such subtask $\tau'$ was avoided in the context of task $\tau$ (equivalent to $t$) in the original plan, and one possible reason for it having been avoided, was that $\tau'$ was prohibited in the context of the task $t$. Lines \ref{alg:addpFTS}--\ref{alg:addpFTF} identify these potential prohibitions, and store them in the temporary variable $\mathit{pF}$. This process is repeated for those states visited as part of the task in Lines \ref{alg:addpFSS}--\ref{alg:addpFSF}.

Finally, the set of potential prohibitions is updated by adding those newly discovered potential prohibitions of $\mathit{pF}$, and removing any prohibitions which definitely do not exist. The set of potential obligations is reduced to those states identified which are definitely obliged; the symbol $\cap_{t}$ is a \emph{context sensitive intersection}. This operation preserves any elements in $\mathit{potO}$ which do not share the same context, and performs a normal intersection operation between any elements in $\mathit{potO}$ and $\mathit{pO}$ which do share context.

\begin{figure}
\begin{center}
\includegraphics[width=0.5\textwidth]{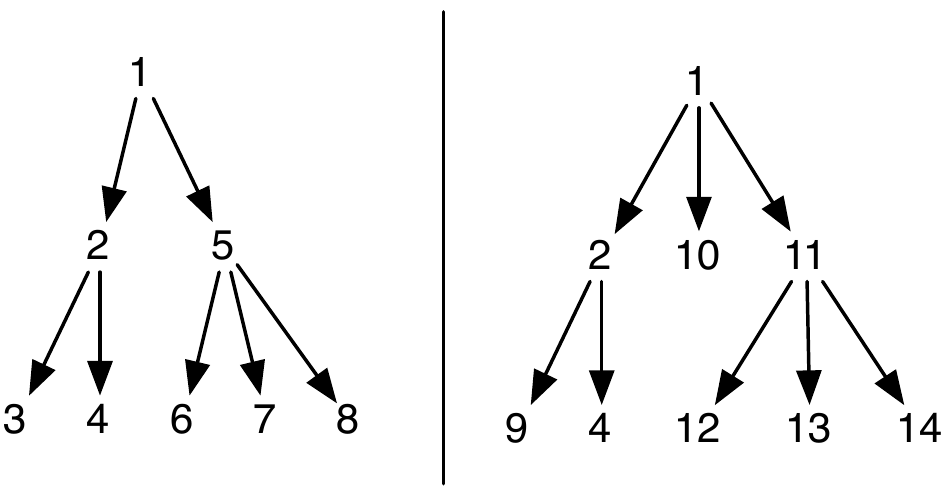}
\end{center}
\caption{\label{fig:networks} An HTN representation of the tasks executed by a plan (left), and an alternative plan (right), to achieve the same goal.}
\end{figure}

Executing this algorithm over a large number of runs of the system will slowly remove from $\mathit{potO}$ those contexts, tasks and states, which are not obliged but were often executed, and will non-monotonically alter the set of prohibitions $\mathit{potF}$.
\begin{example}
As a simple example, assume a run was seen wherein primitive tasks $t_{3},t_{4}, t_{6},t_{7}$ and $t_{8}$ occurred. The plan recogniser reconstructed the plan shown on the left of Figure \ref{fig:networks} for this run, with the alternative plan shown on the right also being a potential plan to execute task $t_{1}$. For clarity, we ignore temporal constraints within these plans, and show only the index of the tasks.

Considering only the plans on the left, our algorithm would generate obligations of the form
$$\begin{array}{l}
\Obl_{t_{1}}t_{2},\Obl_{t_{1}}t_{3}, \Obl_{t_{1}}t_{4}, \ldots \\
\Obl_{t_{2}}t_{3}, \Obl_{t_{2}}t_{4}, \Obl_{t_{5}}t_{6}, \ldots \\
\end{array}
$$
When considering the second tree, some of these obligations would be pruned, and potential prohibitions would also be introduced, resulting in norms of the form
$$\begin{array}{l}
\Obl_{t_{1}}t_{2}, \Obl_{t_{2}}t_{4}, \Obl_{t_{1}}t_{4}, \\
\Frb_{t_{1}}t_{9}, \Frb_{t_{1}}t_{10}, \ldots
\end{array}
$$
\end{example}

\subsection{Allowing Norm Violation}\label{sec:normViol}

Our basic approach is, in some sense, the dual of violation signal based approaches such as \cite{SavarimuthuJASSS2010} --- it can only obtain an accurate model of the norms in the system when no violations take place. Consider for example obligations. Since the set $\mathit{potO}$ shrinks monotonically, a violation of an obligation would permanently remove it from the set of potential obligations. In this section, we describe an extension of our basic approach which aims to overcome this limitation.

\begin{algorithm}[t]
{\small
\begin{algorithmic}[1]
\Require $R$, a set of runs
\Require $\mathit{OT,FT}$, thresholds in $\mathcal{R}^{+}$
\Function{TLearnNorms}{$R$}
  \State initialise $\mathit{OC}$ to $(0,0)$ for every possible obligation.
  \State initialise $\mathit{FC}$ to $(0,0)$ for every possible prohibition.
  \ForAll{$r \in R$}
    \State $(\mathit{OC,FC}) \leftarrow$ \Call{UpdateCounter}{$\mathit{r,OC,FC}$}  
  \EndFor
  \ForAll{($\Obl_{y}z,\mathit{oy,on}) \in \mathit{OC}$} \label{alg:tlnOS}
    \If{($\mathit{on}=0$ and $oy > 0$) or ($\mathit{on} \neq 0$ and $\mathit{oy/on}>\mathit{OT})$}
      \State $\mathit{potO}= \mathit{potO} \cup \{\Obl_{y}z \}$
    \EndIf  
  \EndFor \label{alg:tlnOF}
  \ForAll{($\Frb_{y}z,fy,fn) \in  \mathit{FC}$} \label{alg:tlnFS}
    \If{$fn=0$ or $\mathit{fy/fn}>\mathit{FT}$} \label{alg:tlnF}
      \State $\mathit{potF}=\mathit{potF} \cup  \{ \Frb_{y}z \}$
    \EndIf
  \EndFor \label{alg:tlnFF}
  \ForAll{$y,z$ such that both $\Obl_{y}z$ and $\Frb_{y}z$}
    \State remove $\Obl_{y}z$ from $\mathit{potO}$, $\Frb_{y}z$ from $\mathit{potF}$
  \EndFor
  \State \Return $\mathit{potO,potF}$
\EndFunction
\end{algorithmic}
}
\caption{Threshold-based filtering heuristic}
\label{alg:thresholdLearnNorms}
\end{algorithm}

Algorithm \ref{alg:thresholdLearnNorms} outlines this extension. In turn, it uses another function, \Call{UpdateCounter}{$\mathit{r,OC,FC}$}, which is described in Algorithm \ref{alg:updateCounter}. The latter algorithm takes a single run, and a pair of counters (described next), and updates these counters.

Counters $\mathit{OC}$ and $\mathit{FC}$ respectively store a pair of values for every possible obligation and every possible prohibition that could exist in the system (requiring the same compression techniques discussed in Section \ref{pg:mitigation} when implemented). The first element of this pair is the number of times the obligation (or prohibition) appears to indeed be an obligation (or prohibition). The second element is associated with the number of times the obligation (or prohibition) appears \emph{ not to be} a valid obligation or prohibition.

Lines \ref{alg:tlnOS}--\ref{alg:tlnOF} compute the set of possible obligations by checking whether the ratio between an obligation existing, and not existing, exceeds some threshold \cite{SavarimuthuJASSS2010}\footnote{While we utilise a ratio, an additive approach, or some other technique are equally applicable.}. In doing so, we ensure that there must be at least one situation in which the obligation is a potential obligation.

Lines \ref{alg:tlnFS}--\ref{alg:tlnFF} perform a similar operation for prohibitions. It should however be noted that no positive examples of a prohibition are needed if no negative examples exist in order for the prohibition to be considered a potential prohibition (Line \ref{alg:tlnF}). Finally, we perform a sanity check, removing any prohibitions and obligations with the same context and normative condition.

\begin{algorithm}[t]
{\small
\begin{algorithmic}[1]
\Require $r$, a single run
\Require $FC$ a relation of counter tuples, of the form $\Frb_{y}z \times \mathcal{N} \times \mathcal{N}$
\Require $OC$ a relation of counter tuples, of the form $\Obl_{y}z \times \mathcal{N} \times \mathcal{N}$
\Function{UpdateCounter}{$r$}
  \State $\pi \leftarrow$ \Call{PlanRecogniser}{$r$}
  \ForAll{tasks $t \in \pi$ do}
    \ForAll{subtasks $t' \in \pi$ of $t$}
      \State $(oy,on) \leftarrow \mathit{OC}[\Obl_{t}t']$ \label{alg:ucOIS}
      \State $\mathit{OC}[\Obl_{t}t'] \leftarrow (oy+1,on)$ \label{alg:ucOIF}
      \State $\mathit{(fy,fn)} \leftarrow \mathit{FC}[\Frb_{t}t']$ \label{alg:ucFIS}
      \State $\mathit{FC}[\Frb_{t}t'] \leftarrow (fy,fn+1)$ \label{alg:ucFIF}
    \EndFor
    \ForAll{states $s$ transitioned through as part of $t$} \label{alg:ucSIS}
      \State $\mathit{(oy,on)} \leftarrow \mathit{OC}[\Obl_{t}s]$
      \State $\mathit{OC}[\Obl_{t}s] \leftarrow \mathit{(oy+1,on)}$
      \State $\mathit{(fy,fn)} \leftarrow \mathit{FC}[\Frb_{t}s]$
      \State $\mathit{FC}[\Frb_{t}s] \leftarrow \mathit{(fy,fn+1)}$
    \EndFor  \label{alg:ucSIF} 

    \ForAll{$\pi' \leftarrow$ all possible plans with the same start state as $\pi$ and with a goal identical to $t$} \label{alg:ucP2S}
       \ForAll{tasks $\tau \in \pi' $}
         \ForAll{subtasks $\tau' \in \pi'$ of $\tau$, which are not subtasks of $\tau \in \pi$}
           \State $\mathit{(fy,fn)} \leftarrow  \mathit{FC}[\Frb_{\tau} \tau']$
           \State $\mathit{FC}[\Frb_{\tau} \tau'] \leftarrow \mathit{(fy+1,fn)}$
         \EndFor
         \ForAll{states $\Sigma$ visited as part of $\tau \in \pi'$ and not for $\tau \in \pi$}
           \State $\mathit{(fy,fn)} \leftarrow \mathit{FC}[\Frb_{\tau} \Sigma]$
           \State $\mathit{FC}[\Frb_{\Sigma} \tau'] \leftarrow \mathit{(fy+1,fn)}$
         \EndFor 
       \EndFor    
     \EndFor   \label{alg:ucP2F}
  \EndFor 
  \State \Return $\mathit{OC,FC}$
\EndFunction
\end{algorithmic}
}
\caption{Updating potential norm occurrences.}
\label{alg:updateCounter}
\end{algorithm}

We now turn our attention to the \Call{UpdateCounter}{$\mathit{r,OC,FC}$} function, described in Algorithm \ref{alg:updateCounter}. This algorithm bears similarities to Algorithm \ref{alg:learnNorms}. We begin by utilising the plan recogniser to identify the plan executed in the current run. Again, our algorithm operates in two phases, first considering the current plan, and then considering alternative plans which achieve the same goal. Also, as was previously done, we consider every task found within the plan, and all its subtasks. A  task $t$ and a subtask $t'$ executed as part of the plan may have been executed due to an obligation, and their presence thus increases the likelihood that $\Obl_{t}t'$ is an obligation. We therefore increment the obligation counter to reflect this fact (Lines \ref{alg:ucOIS},\ref{alg:ucOIF}). Similarly, the execution of this task and subtask means that a prohibition on this action does not exist, reducing  its likelihood (Lines \ref{alg:ucFIS},\ref{alg:ucFIF}). Lines \ref{alg:ucSIS}--\ref{alg:ucSIF} repeat this check for tasks and their constituent states.

The second phase  (Lines \ref{alg:ucP2S}--\ref{alg:ucP2F}) again considers all alternative plans, as generated by the planner. Any tasks and subtasks not executed in the plan are potential prohibitions, and their counter is appropriately incremented. This operation is repeated for the visited states in the alternative plan. Finally, the algorithm returns the updated counters.

%

\section{Discussion and Related Work}
\label{sec:discussion}

Savarimuthu and Cranefield \cite{Savarimuthu2011} provide a detailed survey regarding the state of the art in norm learning. 
Existing approaches can be broadly classified into three categories: experiential, observational and communication-based. 
Experiential techniques rely on an agent's perception of sanctions and rewards being applied to its own actions and deriving a model of the norms within the environment from these experiences. 
Observational techniques rely on an agent tracking the behaviour of others and trying to identify norms from the sanctions and rewards applied to others, before acting in the environment. 
Finally, communication-based techniques assume that an agent either asks other agents for the set of norms in force within a society, or is told these norms. 
While our approach is firmly associated with observational techniques within this classification, we drop the assumption that an agent can observe sanctions and rewards being applied to other agents (c.f.~the violation signals of \cite{SavarimuthuJASSS2010}). 

In unpublished work, we modelled a domain consisting of a road network in which agents drive from place to place, but must adhere to certain norms, such as driving on the correct side of the road, avoiding (or passing through) certain areas of the network, and so on. 
We introduced a new agent into this environment and left it to learn the norms by observing the behaviour of pre-existing agents using an implementation of our approach. 
This domain consisted of a hypergraph (to represent multi-lane roads), with agents planning to travel from one node to another using an HTN domain containing a hierarchy of tasks to move the agent about. 
Norms prohibiting or obliging agents to enter specific nodes while travelling were created, as were norms prohibiting certain types of behaviour (such as making left turns). 
Executing our algorithm allowed us to identify the set of prohibited and obliged states and tasks when all agents were fully norm compliant. 
As expected, when only few runs were input to the algorithm, many ``false positive'' obligations were obtained. The norm specification, and associated norm identification algorithms in this evaluation operated over entire plans, eschewing the extra complexity of context, and therefore limiting its applicability to the current work. Indeed, a detailed evaluation of our current approach forms one of our intended avenues of future work.

Since one key component of our approach is a plan recognition algorithm, the quality of the norms identified is strongly dependant on the accuracy of the underlying plan recogniser. While it is desirable to utilise a powerful keyhole plan recogniser such as the one described in~\cite{Avrahami-Zilberbrand2007}, we utilised a simpler technique both in order to simplify the description of our approach, and due to the difficulties inherent in extending this plan recogniser from the propositional to the first order domain. In the future, we aim to employ new algorithms based on recent advances in probabilistic parsing-based plan recognition~\cite{Geib2012,Geib2009}. 

We note that given our algorithm's assumption that every transited state is a potential obligation, as the amount of data (plan runs) supplied to the algorithm increases, the number of potential obligations will increase with the total number of steps of all plans processed. 
This large number of potential obligations can be interpreted in two ways. 
First, they might represent false positives with regards to the formally enforced norms in a society.
On the other hand these potential obligations might represent informal conventions that agents tend to follow within a society, either because they have no other choice (given their plan library), or because these choices have a higher utility.  In future work, we intend to evaluate filters on potential obligations due to agents having no alternative plans to fulfil a goal, as well as to look further into plan utilities. 

It should also be noted that our algorithm can also generate false positives in certain situations. 
Apart from the false positive obligations generated when few runs exist, false positive prohibitions can be generated when real prohibitions make it impossible to reach some states (e.g. if a prohibition exists on visiting states (0,1) and (1,1), and these are the only paths to a state (0,0), then our algorithm will identify (0,0) as prohibited). 
However, since norm compliant agents cannot reach such states, such false positives are not a problem in practice. 

Finally, while Section \ref{sec:normViol} described how normative violations can be dealt with, we have not considered an important class of norms that emerges in such situations, namely contrary-to-duty obligations. Such obligations aim to cause corrective actions to occur within a system, such as obliging an agent to pay reparations for their violations. Contrary-to-duty obligations are context dependent, and are not captured by our structure of norms. Indeed, extending our norm representation to capture rich context conditions (as opposed to just tasks as in the current approach) while still performing plan based norm identification forms one direction of future work which we are pursuing.

We are currently investigating several additional avenues of future work.  
First, we currently assume no interference between agent actions. 
Such interference can cause an agent to prefer one plan over another even in the absence of norms, and  we intend to investigate how such situations can be detected and handled by the mechanism. 
Second, different actions can have different costs for an agent, again creating false positives in our current mechanism. 
Complicating this is the idea that norm violation can also be associated with a cost --- if the cost of the latter is smaller than the benefit of a violating plan, a rational agent can choose to violate a plan. 
We must thus extend our framework to take such costs into account. We also intend to investigate the effectiveness of our approach when only part of a run is visible to the plan recogniser, and to combine our plan recognition based approach with violation signal based techniques.

\section{Conclusions}
\label{sec:conclusion}

In this paper we developed an algorithm for norm identification based on 
a plan recognition based algorithm for norm identification.  
Unlike existing approaches, our mechanism can operate in situations where all agents within the system comply with their norms. However, doing so requires agents to always act in a norm compliant manner.
Since this requirement is unrealistic, we extended our plan recognition approach to cater for violations through counting the number of times obligations and prohibitions potentially do, or do not, exist. This, coupled with threshold functions (in the spirit of \cite{SavarimuthuJASSS2010}) allow our approach to operate in domains where violations can occur.

The use of a plan recognition component to drive norm identification appears promising, and we have highlighted several avenues for future work which we intend to actively pursue.

{\small
\textbf{Acknowledgements:} The authors thank FAPERGS for the financial support within process 3541-2551/12-0 and 12/0808-5 under the ACI and ARD programs.
}

\bibliographystyle{abbrv}
\bibliography{felipe}

\begin{thebibliography}{10}

\bibitem{Aldewereld2006}
H.~Aldewereld, F.~Dignum, A.~Garc\'{\i}a-Camino, P.~Noriega, J.~A.
  Rodr\'{\i}guez-Aguilar, and C.~Sierra.
\newblock Operationalisation of norms for usage in electronic institutions.
\newblock In {\em Proc.~of the Fifth Int'l Joint Conf.~on Autonomous Agents and
  Multiagent Systems}, pages 223--225, 2006.

\bibitem{Andrighetto2007}
G.~Andrighetto, R.~Conte, P.~Turrini, and M.~Paolucci.
\newblock Emergence in the loop: Simulating the two way dynamics of norm
  innovation.
\newblock In G.~Boella, L.~W.~N. van~der Torre, and H.~Verhagen, editors, {\em
  Normative Multi-agent Systems}, volume 07122 of {\em Dagstuhl Seminar
  Proceedings}, 2007.

\bibitem{Armentano2007}
M.~G. Armentano and A.~Amandi.
\newblock Plan recognition for interface agents.
\newblock {\em Artif. Intell. Rev.}, 28(2):131--162, 2007.

\bibitem{Avrahami-Zilberbrand2007}
D.~Avrahami-Zilberbrand and G.~A. Kaminka.
\newblock Incorporating observer biases in keyhole plan recognition
  (efficiently!).
\newblock In {\em Proc. AAAI}, 2007.

\bibitem{Erol1994}
K.~Erol, J.~Hendler, and D.~S. Nau.
\newblock {HTN} planning: Complexity and expressivity.
\newblock In {\em Proceedings of the Twelfth National Conference on Artificial
  Intelligence}, volume~2, pages 1123--1128, Seattle, Washington, USA, 1994.
  AAAI Press/MIT Press.

\bibitem{Geib2012}
C.~Geib.
\newblock Considering state in plan recognition with lexicalized grammars.
\newblock 2012.

\bibitem{Geib2009}
C.~W. Geib and R.~P. Goldman.
\newblock A probabilistic plan recognition algorithm based on plan tree
  grammars.
\newblock {\em Artif. Intell.}, 173(11):1101--1132, July 2009.

\bibitem{Geib2007}
C.~W. Geib and M.~Steedman.
\newblock On natural language processing and plan recognition.
\newblock In {\em Proc. 20th Int'l Joint Conf.~on Artif.~Intell.}, pages
  1612--1617, 2007.

\bibitem{Ghallab2004}
M.~Ghallab, D.~Nau, and P.~Traverso.
\newblock {\em Automated Planning: Theory and Practice}.
\newblock Elsevier, 2004.

\bibitem{oren2008}
N.~Oren, S.~Panagiotidi, J.~Vazquez-Salceda, S.~Modgil, M.~Luck, and S.~Miles.
\newblock Towards a formalisation of electronic contracting environments.
\newblock In {\em Proc.~of Coordination, Organization, Institutions and Norms
  in Agent Systems, the Int'l Workshop at AAAI 2008}, pages 61--68, Chicago,
  Illinois, USA, 2008.

\bibitem{Rao1996}
A.~S. Rao.
\newblock {A}gent{S}peak({L}): {BDI} agents speak out in a logical computable
  language.
\newblock In W.~V. de~Velde and J.~W. Perram, editors, {\em Proc.~of the 7th
  MAAMAW}, volume 1038 of {\em LNCS}, pages 42--55. Springer-Verlag, 1996.

\bibitem{Savarimuthu2011}
B.~T.~R. Savarimuthu and S.~Cranefield.
\newblock Norm creation, spreading and emergence: A survey of simulation models
  of norms in multi-agent systems.
\newblock {\em Multiagent and Grid Systems}, 7(1):21--54, 2011.

\bibitem{Savarimuthu2010a}
B.~T.~R. Savarimuthu, S.~Cranefield, M.~Purvis, and M.~K. Purvis.
\newblock Identifying conditional norms in multi-agent societies.
\newblock In M.~D. Vos, N.~Fornara, J.~V. Pitt, and G.~A. Vouros, editors, {\em
  COIN@AAMAS{\&}MALLOW}, volume 6541 of {\em Lecture Notes in Computer
  Science}, pages 285--302. Springer, 2010.

\bibitem{Savarimuthu2009}
B.~T.~R. Savarimuthu, S.~Cranefield, M.~A. Purvis, and M.~K. Purvis.
\newblock Internal agent architecture for norm identification.
\newblock In {\em Proc.~of the 5th Int'l Workshop on Coordination,
  organizations, institutions, and norms in agent systems}, COIN'09, pages
  241--256, Berlin, Heidelberg, 2010. Springer-Verlag.

\bibitem{SavarimuthuJASSS2010}
B.~T.~R. Savarimuthu, S.~Cranefield, M.~A. Purvis, and M.~K. Purvis.
\newblock Obligation norm identification in agent societies.
\newblock {\em Journal of Artificial Societies and Social Simulation}, 13(4),
  2010.

\bibitem{Sukthankar2007}
G.~Sukthankar.
\newblock {\em Activity Recognition for Agent Teams}.
\newblock PhD thesis, Carnegie Mellon University, 2007.

\end{thebibliography}

\end{document}